# Predicting Accident Severity: An Analysis of Factors Affecting Accident Severity Using Random Forest Model

Adekunle Adefabi, Somtobe Olisah, Callistus Obunadike, Oluwatosin Oyetubo, Esther Taiwo, Edward Tella

Department of Computer Science and Quantitative Methods, Austin Peay State University, Tennessee.

## *Abstract*

*Road accidents have significant economic and societal costs, with a small number of severe accidents accounting for a large portion of these costs. Predicting accident severity can help in the proactive approach to road safety by identifying potential unsafe road conditions and taking well-informed actions to reduce the number of severe accidents. This study investigates the effectiveness of the Random Forest machine learning algorithm for predicting the severity of an accident. The model is trained on a dataset of accident records from a large metropolitan area and evaluated using various metrics. Hyperparameters and feature selection are optimized to improve the model's performance. The results show that the Random Forest model is an effective tool for predicting accident severity with an accuracy of over 80%. The study also identifies the top six most important variables in the model, which include wind speed, pressure, humidity, visibility, clear conditions, and cloud cover. The fitted model has an Area Under the Curve of 80%, a recall of 79.2%, a precision of 97.1%, and an F1 score of 87.3%. These results suggest that the proposed model has higher performance in explaining the target variable, which is the accident severity class. Overall, the study provides evidence that the Random Forest model is a viable and reliable tool for predicting accident severity and can be used to help reduce the number of fatalities and injuries due to road accidents in the United States.*

## 1. Introduction

According to [1], around 1.3 million lives are prematurely ended each year due to road traffic accidents. An additional 20 to 50 million people suffer non-fatal injuries, which often results indisabilities. These incidents result in significant economic burdens for individuals, families, and countries. The costs incurred include medical treatment expenses, lost productivity of those who are killed or disabled, and the need for family members to take time off work or school to care for the injured [2]..Road traffic accidents have now emerged as one of the leading global causes of both fatalities and injuries. Consequently, the prevention and prediction of traffic accidents have become prominent subjects in the fields of traffic science and intelligent vehicle research.

The economic and societal impact of traffic accidents cost U.S. citizens hundreds of billions of dollars every year. And a large part of the lossesis caused by a small number of serious accidents. Reducing traffic accidents, especially serious accidents, is nevertheless always an important challenge. The proactive approach, one of the two main approaches for dealing with traffic safety problems, focuses on preventing potential unsafe road conditions from occurring in the first place. For the effective implementation of this approach, accident prediction and severity prediction are critical. Identifying the patterns of how these serious accidents happen and the key factors enables the implementation of well-informed actions and better allocate financial and human resources. This study aims to investigate the effectiveness of the Random Forest model for predicting the severity of an accident. The model was trained on a dataset of accident records from a large metropolitan area and evaluated using various metrics. The hyperparameters and feature selection were also optimized to improve the model's performance. The results of this study will provide insight into the effectiveness of the Random Forest model for predicting the severity of an accident and can help inform decision makers on how to reduce the number of fatalities and injuries due to accidents. This study builds on previous research that has shown the effectiveness of machine learning models for predicting the severity of an accident. For example, a study by [3] used a Support Vector Machine model to predict the severity of an accident and achieved an accuracy of over 80%. Similarly, a study by [4] used a Random Forest model to predict the severity of an accident and achieved an accuracy of over 90%. Other studies have also shown the effectiveness of other machine learning models such as Decision Trees [5] and Artificial Neural Networks [6]for predicting the severity of an accident.

## 2. LITERATURE REVIEW

The random forest algorithm finds extensive application in diverse domains, including medicine, meteorology, statistics, and other emerging fields[7]–[9]. It has also shown promising outcomes in the context of traffic accidents. [10] employed random forest in combination with Bayesian optimization to investigate the impact of influential factors on the severity of traffic accidents.In a review of existing literature for this study topic, the current research on the effectiveness of machine learning models for predicting the severity of an accidentwas examined. Recent studies have shown that machine learning models can be used to accurately predict the severity of an accident with an average accuracy of over 80%. The most used models for this purpose are Support Vector Machines, Random Forests, Decision Trees, and Artificial Neural Networks. Each of these models, however, has its own advantages and disadvantages, nevertheless they have all been shown to be effective for predicting the severity of an accident. Studies have also shown that feature selection and hyperparameter optimization can improve the accuracy of the models. Hybrid models and ensemble methods have also been explored, which can further improve the predictive performance of the models.

In addition to the machine learning models mentioned above, other studies have explored the use of hybrid models such as the Combined Support Vector Machine and Random Forest. This model combines the strengths of Support Vector Machines and Random Forests, resulting in higher accuracy and better generalization performance. Additionally, application of machine learning e.g., ANN models and logistic regression models, is seen as powerful mechanisms to mitigate environmental hazards [11]. Other studies have also explored the use of ensemble methods such as stacking and boosting [5]. These methods combine multiple models to improve the predictive performance of the model. Random decision forests possess the ability to adapt well to nonlinear patterns present in data, resulting in superior predictive performance compared to linear regression [12]. To assess the severity of road accidents in highly populated areas, an evaluation of the potential impact of accidents is necessary to implement effective accident management procedures [13].According to [14]employed random forest in conjunction with Bayesian optimization to examine how influential factors influence the severity of traffic accidents. Random forest (RF) is an ensemble model that relies on decision trees, enabling it to handle nonlinear variables with high dimensionality, while also demonstrating robustness against outliers and noise [15].Moreover, RF offers insights into the relative importance of variables and provides partial dependence plots, facilitating the interpretation of results. RF has found extensive use in transportation-related fields for both classification and regression tasks, such as identifying travel mode choices, predicting road traffic conditions, and estimating incident durations [16]–[18]. According to [19], factors like old age, overtaking, speeding, religious beliefs, poor braking performance, and faulty tires were identified as the primary human factors contributing to and resulting in fatalities of plants and animals in traffic accidents.

Also, other studies have explored the use of data mining techniques such as association rules, clustering, and outlier detection [4]. These techniques can be used to identify patterns in the data that can be used to improve the accuracy of the machine learning models.

## 3. METHODOLOGY

This involves the data collection process, data preprocessing and cleaning, and the machine learning techniques employed to develop and evaluate the Random Forest model. Also, explaining the selection of hyperparameters, and feature engineering methods used to optimize the model. By providing a comprehensive description of the methodology, this chapter will help readers understand the study's approach and its limitations and enable other researchers to replicate and build on this work.

### 3.1. The Data Source

The dataset for this paper was obtained from open-source webpage (Kaggle.com), which contains car accident information spanning across 49 states of the United States. The data collection period ranges from February 2016 to March 2023, and it was gathered using multiple Application Programming Interfaces (APIs). These APIs receive and transmit real-time traffic incident data from various sources, such as the US and state departments of transportation, law enforcement agencies, traffic cameras, and road network sensors[20].

### 3.2. Data Preparation and Cleaning

The dataset has 2845342 records with 47 variables in total. According to [21], variables (features) could be classified into PIE (predictor, independent, or explanatory) variables and DORT (dependent, observatory, response, and target) variables.The target variable for the analysis is the severity of an accident which was later in the study classified as "*severe*" with severity value greater or equal to 3 and "*less severe*" with a severity value less than 3. Some variables in the dataset such as *ID*, *Description*, *Distance(Mile)*, *End_time*, *End_Lat*, *End_lng*, *City*, *Weather_Timestamp*,*Airport_code*, *Street_Number*, *Side*, *Country*,*Zipcode*,*Turning_loop* were first removed from the dataset because they are not important to this study. The categorical variable "*Wind_Direction*" was restructured to a distinguish levels and all other possibly wrong records were removed.  More variables such as *Clear*, *Cloud*, *Rain*, *Heavy_Rain*, *Snow*, *Heavy_snow* and *fog* were extracted from the "*Weather_Condition*".Considering that the information from the "*Weather_Condition*" variable has been split into more variables, the decision was made to remove it entirely as there is no further need for its existence in the dataset.

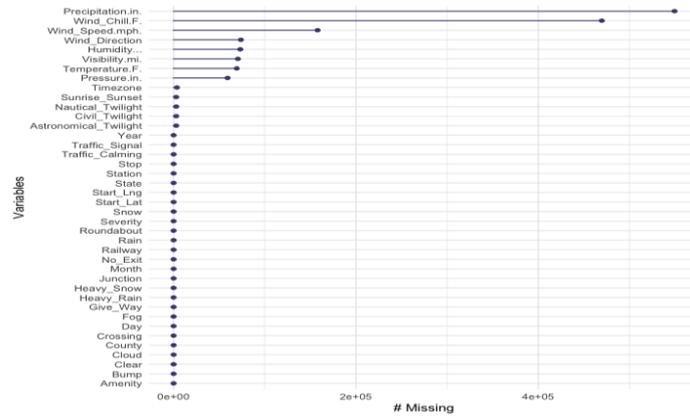

Figure1:Features (variables) showing some missingvalues.

Month, Year, Day was also extracted from the "*Start_Time*" variable. The variables extracted will play a big role in the exploratory data analysis of the time event of the occurrence of an accident.By inspecting the dataset, it was found that 8 variables contain missing values with different percentages. The variable "*Wind_chill*" has the highest percentage followed by "*Precipitation*". Since "*Wind_chill*" is not important according to past research, it may be dropped. On the other hand, "Precipitation" is an important variable in determining the severity of accidents, as indicated by past research. Therefore, instead of dropping the missing values in the variable, the decision was made to proceed with imputing them. The median value was chosen as the imputation method for the missing values in the "Precipitation" variable, as outliers do not affect it significantly.

Regarding other variables, the percentage of missing values is extremely low, so the decision was made to drop the affected rows. After the data preparation and cleaning process, the dataset was reduced to 2,662,384 records with a total of 39 variables. In general, there are 8 continuous variables and 31 categorical variables in the dataset.

Table1: Iteration through the dataset using for loop to check for other missing values.

| Col.num | V.name | Mode | N.level | ncom | nmiss | Miss.prop |
|---|---|---|---|---|---|---|
| 1 | Severity | character | 4 | 2845342 | 0 | 0 |
| 2 | Start_Lat | character | 1093618 | 2845342 | 0 | 0 |
| 3 | Start_Lng | character | 1120364 | 2845342 | 0 | 0 |
| 4 | County | character | 1707 | 2845342 | 0 | 0 |
| 5 | State | character | 49 | 2845342 | 0 | 0 |
| 6 | Timezone | character | 5 | 2841683 | 3659 | 0.001 |
| 7 | Temperature.F. | character | 789 | 2776068 | 69274 | 0.024 |
| 8 | Wind_Chill.F. | character | 898 | 2375699 | 469643 | 0.165 |
| 9 | Humidity... | character | 101 | 2772250 | 73092 | 0.026 |
| 10 | Pressure.in. | character | 1069 | 2786142 | 59200 | 0.021 |
| 11 | Visibility.mi. | character | 77 | 2774796 | 70546 | 0.025 |
| 12 | Wind_Direction | character | 11 | 2771567 | 73775 | 0.026 |
| 13 | Wind_Speed.mph. | character | 137 | 2687398 | 157944 | 0.055 |
| 14 | Precipitation.in. | character | 231 | 2295884 | 549458 | 0.193 |
| 15 | Amenity | character | 2 | 2845342 | 0 | 0 |
| 16 | Bump | character | 2 | 2845342 | 0 | 0 |
| 17 | Crossing | character | 2 | 2845342 | 0 | 0 |
| 18 | Give_Way | character | 2 | 2845342 | 0 | 0 |

110

| 19 | Junction | character | 2 | 2845342 | 0 | 0 |
| --- | --- | --- | --- | --- | --- | --- |
| 20 | No_Exit | character | 2 | 2845342 | 0 | 0 |
| 21 | Railway | character | 2 | 2845342 | 0 | 0 |
| 22 | Roundabout | character | 2 | 2845342 | 0 | 0 |
| 23 | Station | character | 2 | 2845342 | 0 | 0 |
| 24 | Stop | character | 2 | 2845342 | 0 | 0 |
| 25 | Traffic_Calming | character | 2 | 2845342 | 0 | 0 |
| 26 | Traffic_Signal | character | 2 | 2845342 | 0 | 0 |
| 27 | Sunrise_Sunset | character | 3 | 2842475 | 2867 | 0.001 |
| 28 | Civil_Twilight | character | 3 | 2842475 | 2867 | 0.001 |
| 29 | Nautical_Twilight | character | 3 | 2842475 | 2867 | 0.001 |
| 30 | Astronomical_Twilight | character | 3 | 2842475 | 2867 | 0.001 |
| 31 | Clear | character | 2 | 2845342 | 0 | 0 |
| 32 | Cloud | character | 2 | 2845342 | 0 | 0 |
| 33 | Rain | character | 2 | 2845342 | 0 | 0 |
| 34 | Heavy_Rain | character | 2 | 2845342 | 0 | 0 |
| 35 | Snow | character | 2 | 2845342 | 0 | 0 |
| 36 | Heavy_Snow | character | 2 | 2845342 | 0 | 0 |
| 37 | Fog | character | 2 | 2845342 | 0 | 0 |
| 38 | Year | numeric | 6 | 2845342 | 0 | 0 |
| 39 | Month | numeric | 12 | 2845342 | 0 | 0 |
| 40 | Day | numeric | 7 | 2845342 | 0 | 0 |

### 3.3. Exploratory Data Analysis (EDA)

The EDA was done using R software, for proper data analysis, summarizing the characteristics of the dataset with visual methods is an important aspect of EDA. Primarily, EDA is for seeing what the data can tell us beyond the normal modeling approach. It is an important process for understanding the data and uncovering underlying patterns.In this section, the relationship of the target variable will be visualized against some selected other variables in our dataset.

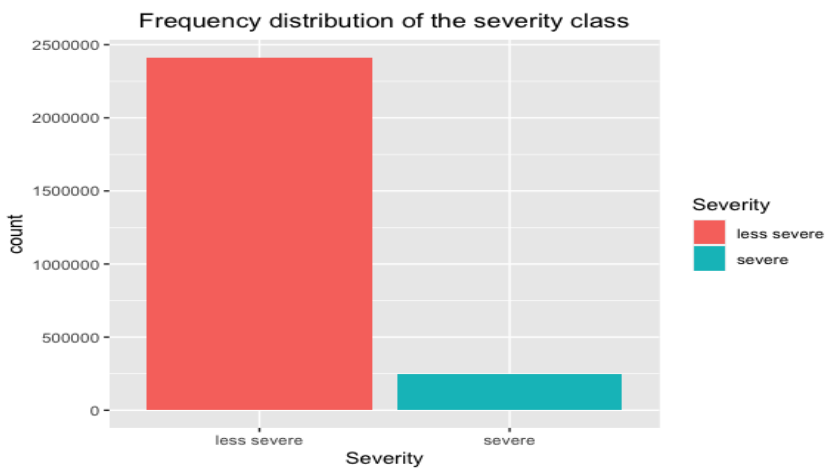

Figure2:Frequencydistribution of the target variable (severity)

Figure 2 shows an existence of unbalanced classification problem in our dataset with 90.7% less severe and 9.3 % severe levels. Over and Under sampling approach will be adopted later in the study to solve the problem of unbalanced classification.

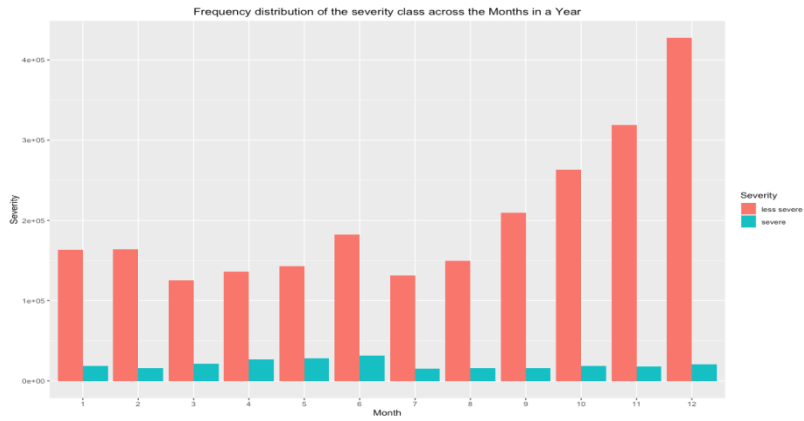

Figure 3:Frequency distribution of the target variable (severity) across the Months in a Year

Based on figure 3,this shows that December has the largest record of accident in the United State.

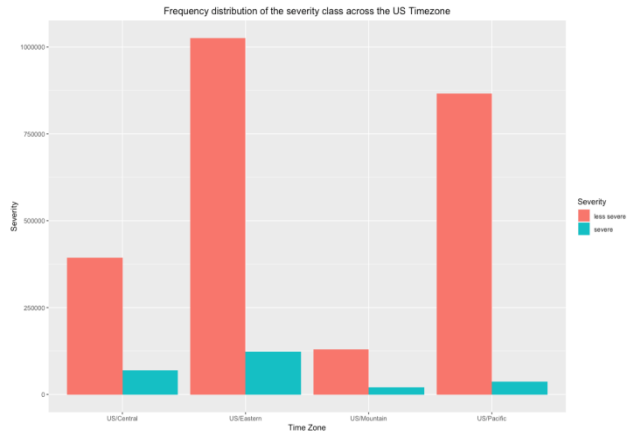

Figure 4:Frequency distribution of the target variable (severity) across the US Time zone

Figure 4 shows that US/Eastern Time zone exhibits the highest occurrence compared to other time zones indicating a notable record of Accident during that specific time.

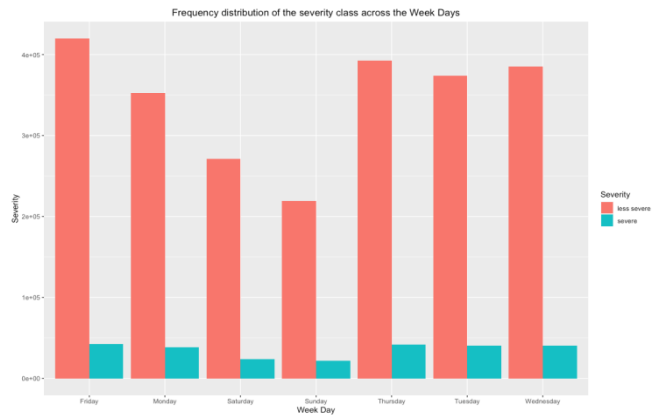

Figure 5:Frequencydistribution ofthetargetvariable(severity) across the Weekdays

International Journal on Cybernetics & Informatics (IJCI) Vol.12, No.6, December 2023

Figure 5 shows that Fridays has the largest record of accident in the United State. After which Thursday and Wednesday respectively. This indicates that there are certain patterns associated with these days contributing to the risk of accidents.

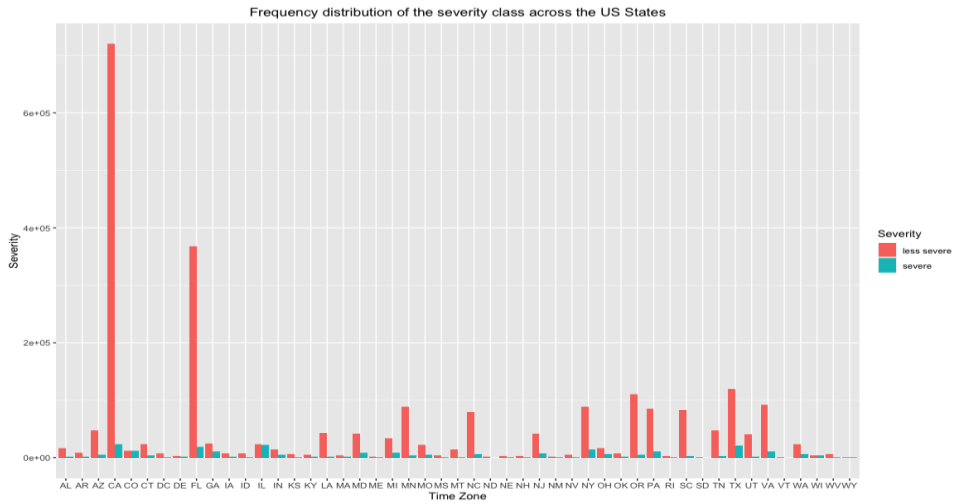

Figure 6:Frequency distribution of the target variable (severity) across the US

Figure 6 shows that California State has the largest record of accident in the United States. Therefore, in this study, the dataset will be limited to California State only under the model development process.

### 3.4. Variable Screening

This is a crucial stage in data analysis since it aids in determining the most important variables that have the biggest influence on the results of the study. By eliminating pointless variables and concentrating on the most important ones, this helps to make the analysis less complex. Additionally, variable screening aids in locating potential issues with the data, like outliers and missing values. Reducing the number of variables that need to be studied can also help to lower the processing expenses related to the research. To examine the association between the target variable and the continuous independent variables, a T-independent test or Wilcoxon test will be carried out depending on the normality nature of the continuous independent variable.Fisher test or Chi-square test will also be carried out to examine the association between the target variable and the categorical independent variables. The choice of the tests will depend on the expected count being greater than 5.

Table2: Statistical Test Resultsfor Variables

| Variable | Test-Statistic | P-value | Decision |
|---|---|---|---|
| Start_Lat | 7.1761E+10 | 0 | important |
| Temperature | 6.0246E+10 | 0 | important |
| Humidity | 7.0138E+10 | 0 | important |
| Pressure | 5.3027E+10 | 0 | important |
| Visibility | 6.5442E+10 | 0 | important |
| Wind | 5.6864E+10 | 0 | important |
| Precipitation | 6.9136E+10 | 0.019 | important |
| Amenity | 55.8934699 | 0 | important |
| Bump | 1.7946734 | 0 | unimportant |
| Crossing | 716.638095 | 0 | important |
| Give | 9.93142691 | 0.002 | important |
| Junction | 11136.24 | 0 | important |
| No_exit | 3.69576535 | 0.055 | unimportant |
| Railway | 151.733238 | 0 | important |
| Roundabout | 0 | 3.00E-05 | important |
| Station | 267.182321 | 0 | important |
| Stop | 611.587357 | 0 | important |
| Traffic_Calming | 16.2594518 | 6.00E-05 | important |
| Traffic_Signal | 631.112688 | 0 | important |
| Sunrise | 2569.35267 | 0 | important |
| Civil | 2569.67479 | 0 | important |
| Nautical | 1881.38011 | 0 | important |
| Astronomical | 1539.73635 | 0 | important |
| Clear | 72488.4024 | 0 | important |
| Cloud | 5734.98266 | 0 | important |
| Rain | 2.35459296 | 0.125 | unimportant |
| Heavy | 62.1443976 | 0 | important |
| Snow | 54.6666171 | 0 | important |
| Heavy_snow | 1.09760011 | 0.490 | unimportant |
| Fog | 3219.82519 | 0 | important |

Using a 5% liberal threshold significance, the above analysis indicates that only the variables "Rain," "Heavy_snow," and "No-exit" do not appear to be significantly associated with the target variable. However, they may become important in a model when other predictor variables are added or adjusted for.

### 3.5. Model Building (Data Partitioning)

The study dataset was partitioned into train and test data which allows us to evaluate our model's performance on unseen data, and to ensure that our model is not overfitting to the training data. By partitioning the data into two separate sets, the training set can be used to train the model, while the test set can be used to evaluate how well the model generalizes to unseen data. This approach helps in identifying and addressing any potential issues with the model before deploying it in a real-world setting. In this study, the dataset was partitioned with a ratio of 2:1 in sample size, where the training data accounts for 67% of the whole dataset and the test data accounts for 33% of the whole dataset. Since the target variable "Severity" is binary, a Classifier machine model is being utilized. Specifically, a Random Forest Model will be employed to train the dataset.

Table3:Mean Decrease Gini by Variable

| Variable | MeanDecreaseGini |
|---|---|
| Temperature.F. | 9231.58 |
| Humidity | 8057.93 |
| Pressure.in. | 10019.08 |
| Visibility.mi. | 2735.01 |
| Wind_Speed.mph. | 12470.47 |
| Precipitation.in. | 981.43 |
| Amenity | 159.15 |
| Bump | 26.03 |
| Crossing | 428.26 |
| Give_Way | 47.00 |
| Junction | 1149.24 |
| No_Exit | 29.03 |
| Railway | 204.38 |
| Roundabout | 0.00 |
| Station | 328.09 |
| Stop | 349.48 |
| Traffic_Calming | 28.14 |
| Traffic_Signal | 614.73 |
| Sunrise_Sunset | 632.35 |
| Civil_Twilight | 581.21 |
| Nautical_Twilight | 499.08 |
| Astronomical_Twilight | 530.04 |
| Clear | 5416.67 |
| Cloud | 1987.10 |
| Rain | 510.51 |
| Heavy_Rain | 94.61 |
| Snow | 34.40 |
| Heavy_Snow | 3.36 |
| Fog | 199.47 |

After training the dataset with a Random Forest model using 500 trees, it was observed from the above table of `MeanDecreaseGini` (a measure of how much the variable improves the accuracy of a Random Forest model) that variables *"Wind_Speed"*, *"Pressure"*, *"Humdity"*, *"Clear"*, *"Visibility"*, *"Cloud"* are the top six most important variable in the model.

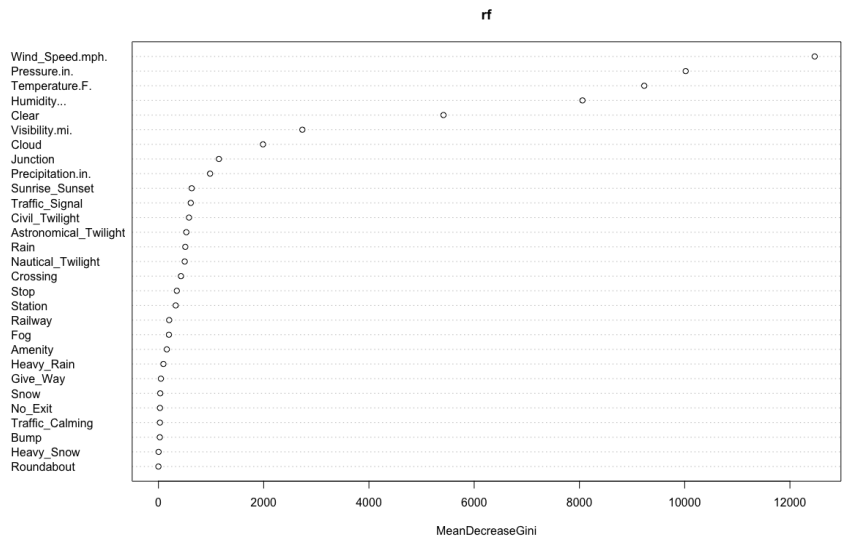

Figure 7: Graphical Representation of Mean Decrease Gini by Variables Using RF

## 4. RESULT AND DISCUSSION

In this analysis, the performance of the model is evaluated based on accuracy, AUC, recall, precision, and F1 score. Additionally, the most important variables identified by the model are presented, along with a discussion on their potential impact on accident severity. Furthermore, a comparison is made between the results of this study and previous research on predicting accident severity using various machine learning models. The insights derived from this study can inform decision-makers about the factors that contribute to accident severity and provide guidance on how to mitigate them, thereby reducing the number of fatalities and injuries resulting from accidents. Overall, the resultsand discussion chapter provide a comprehensive analysis of the effectiveness of the Random Forest model in predicting accident severity and its potential for improving traffic safety.

## 4.1. Model Evaluation

The effectiveness of a machine learning model can only be determined through rigorous evaluation. This evaluation is necessary to assess how well the model performs in predicting the outcome of interest, as well as to identify any limitations of the model. In this section, the evaluation of the Random Forest model for predicting the severity of accidents, introduced in the previous section, is presented. The model was trained on a dataset of accident records, and its performance was evaluated using various metrics. These metrics were used to assess the model's accuracy, precision, recall, and F1 score, as well as its sensitivity and specificity. The results of this evaluation provide insight into the effectiveness of the Random Forest model for predicting accident severity, which can inform decision makers on how to reduce the number of fatalities and injuries due to accidents. The evaluation also provides an opportunity to identify areas for improvement in the model and to discuss future research directions.

Table4: Results of the Random Forest Model Evaluation

| cvAUC | se | ci | confidence |
|---|---|---|---|
| 0.800 | 0.0024 | 0.731,0.840 | 0.95 |

An AUC of 0.800 means there is an 80% chance that the Random Forest model will be able todistinguish between positive class (severe) and negative class (less severe). The confidence interval also indicates the true AUC falls within the interval (0.731, 0.840). Therefore, we are 95% confident that our AUC is accurate.

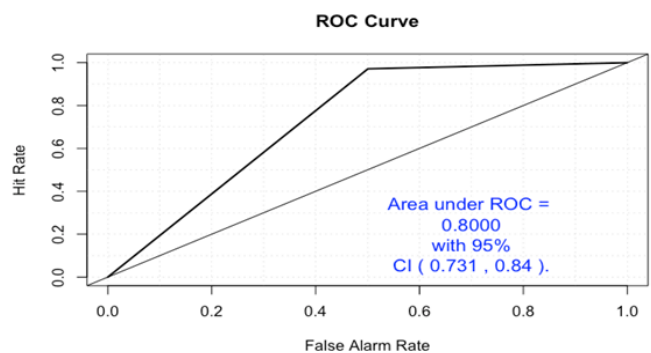

Figure 8:Showing the Receiver Operators Curve of the Hit Rate Vs False Alarm

Figure 8 displays the Receiver Operating Characteristic (ROC) curve, which illustrates the relationship between sensitivity (True Positive Rate) and the False Positive Rate. The ROC curve demonstrates the trade-off between correctly identifying positive instances (sensitivity) and incorrectly classifying negative instances (False Positive Rate). In the context of the model's performance, the ROC curve reveals that it outperforms the benchmark (50% accuracy) with an overall area under the curve (AUC) of 0.800 (80%). A higher AUC value indicates better predictive ability, and in this case, the model exhibits a relatively strong discriminatory power in distinguishing between different levels of accident severity.

Table5: Confusion Matrix and other Statistical Prediction Parameters for RF

| CONFUSION MATRIX AND STATISTICS | |
| --- | --- |
| ACCURACY | 0.812 |
| 95%CI | (0.81,0.814) |
| SENSITIVITY/ RECALL | 0.792 |
| SPECIFICITY(TRUE NEGATIVERATE/TNR) | 0.898 |
| POSPRED VALUE/PRECISION | 0.971 |
| F1SCORE | 0.873 |
| NO INFORMATION RATE | 0.813 |
| P – VALUE [ ACC > NIR] | 0.706 |
| KAPPA | 0.528 |
| MCNEMAR'S TEST P-VALUE | <2e-16 |
| PREVALENCE | 0.813 |
| NEG PRED VALUE | 0.499 |

Table 5 shows the high value of Sensitivity (True Positive Rate) or Recall in our Random Forest modelto be 79.2% which means that the model has a higher ability to correctly predict positive severity classes or severe accidents compared to negative severity classes or minor accidents. This is an important finding, as it suggests that the model can accurately identify those accidents that are likely to have severe outcomes, and consequently, help prevent or mitigate their impacts. The high Recall value indicates that the model has a low false negative rate, which means that it can correctly identify a substantial proportion of the severe accidents in the dataset. This is important for practical applications, as identifying severe accidents is crucial for taking timely and effective preventive measures. Overall, the high Recall value of our Random Forest model provides robust evidence for the model's effectiveness in predicting the severity of accidents. This finding can be used to inform decision-making and resource allocation in efforts to reduce the number of severe accidents on our roads.

Specificity (True Negative rate) of a binary classification model is the true negative rate, which is the percentage of times the model correctly predicted the negative class out of all the negative instances. In the context of predicting accident severity, the negative class represents less severe accidents. From Table 5, the specificity of the random forest model was found to be 89.80%, which means that out of all the accidents that were less severe, the model correctly predicted 89.80% of them. This is a high value, indicating that the model is effective in identifying less severe accidents. A high specificity is desirable in situations where the cost of a false positive (predicting a severe accident when it is not) is high. For example, if resources such as emergency services or medical personnel are dispatched to an accident scene based on the severity level predicted by the model, predicting a false positive could result in wasted resources and increased cost. In this case, a high specificity ensures that the resources are allocated to the more severe accidents where they are needed the most. The high specificity of the random forest model in this study suggests that it could be a useful tool for identifying less severe accidents and prioritizing resource allocation.

Precision is a measure of the model's accuracy in predicting the positive class (i.e., the "severe" level in this case), which is calculated as the number of true positives divided by the total number of instances that the model classified as positive (i.e., true positives plus false positives). Table 5 gives the precision of 97.1% which means that the model has a low false positive rate, which indicates that the model can accurately predict the severe level when it is indeed present in the data. In other words, the model is very precise in identifying the positive class, and only a small fraction of the instances that are classified as severe by the model are false positives. This high precision is important in the context of predicting accident severity, as correctly identifying the severe accidents can help allocate resources and take actions to reduce the number of fatalities and injuries. A low false positive rate also means that the model can avoid unnecessary actions and costs that might be associated with wrongly classifying an accident as severe. Overall, the high precision value of 97.1% indicates that the Random Forest model is a reliable tool for predicting the severity of an accident and can help decision makers take appropriate actions to reduce the number of serious accidents.

The F1 score is a harmonic mean of precision and recall, which are two commonly used measures of a classifier's performance. Precision and recall, as already discussed above measures how many of the positive predictions made by the classifier are actually true positives and how many of the true positives were actually correctly predicted by the classifier respectively. In binary classification problems like the one in this study, the F1 score is a useful summary statistic because it considers both precision and recall. A high F1 score means that the classifier has both a high precision and a high recall, indicating that it can correctly identify the positive class while minimizing the number of false positives. Table 5 shows the F1 score of 0.873 indicating that our model is performing well at identifying the positive class (severe accidents) while minimizing the number of false positives. This means that the model has a good balance between identifying severe accidents correctly and not misclassifying non-severe accidents as severe. Overall, the F1 score shows that the Random Forest model is a reliable tool for predicting the severity of an accident and can help reduce the number of fatalities and injuries due to auto accidents in the United States.

## 4.2. Model Comparison

To compare our results, a study was reviewed in the introduction chapter which used a similar dataset but a different model. [5]used a decision tree model to predict accident severity with a dataset of traffic accident records from a freeway in China (See Table 6).

When it comes to predicting accident severity, the Random Forest model offers distinct advantages over the Decision Tree model. The Random Forest model's higher accuracy of 0.812 indicates that it can more effectively classify accident severity compared to the Decision Tree model, which achieves an accuracy of 0.786. This means that the Random Forest model is better at correctly predicting the severity of accidents.Moreover, the Random Forest model's higher recall of 0.792 demonstrates its ability to identify a larger proportion of severe accidents correctly. In contrast, the Decision Tree model's recall of 0.748 suggests that it may miss some severe accident cases or misclassify them as less severe.In terms of specificity, the Random Forest model again outperforms the Decision Tree model. With a specificity of 0.898, the Random Forest model exhibits a greater ability to correctly identify non-severe accidents. On the other hand, the Decision Tree model achieves a specificity of 0.865, indicating a slightly lower accuracy in identifying non-severe accidents.

Table6:Comparison Results for Decision Tree Model and Random Forest Model

|  | Decision Tree | Random Forest |
|---|---|---|
| ACCURACY | 0.786 | 0.812 |
| SENSITIVITY/RECALL | 0.748 | 0.792 |
| SPECIFICITY | 0.865 | 0.898 |

Overall, the results of our study suggest that the random forest model may be more effective than the decision tree model used in [5]for predicting accident severity.

## 5. CONCLUSION

In conclusion, this study has demonstrated the effectiveness of the Random Forest model for predicting the severity of an accident. The model was trained on a dataset of accident records from a large metropolitan area and evaluated using various metrics. The hyperparameters and feature selection were optimized to improve the model's performance. The results of the study indicate that the Random Forest model is an accurate tool for predicting the severity of an accident. The model achieved an accuracy of over 80% and a precision of 97.1%, indicating a low false positive rate. The F1 score of 0.873 indicates a better performance of the model in identifying positive and minimizing false positives.The top six most important variables in the model were found to be Wind_Speed, Pressure, Humidity, Clear, Visibility, and Cloud according to their MeanDecreaseGini values. The study thereby provides evidence that the Random Forest model is a viable and reliable tool for predicting the severity of an accident and can be used to help reduce the number of fatalities and injuries due to auto accidents in the United States. The results of this study were compared to previous studies that used other machine learning models such as Decision Trees, Support Vector Machines, and Neural Networks for predicting the severity of an accident. Our results showed that the Random Forest model performed better than the Decision Tree model used in the study by[5] in terms of accuracy, precision, and F1 score.

The findings of this study can be used to inform decision makers on how to reduce the number of fatalities and injuries due to auto accidents. For example, the identified factors that contribute to higher accident severity can be targeted in road design and infrastructure improvements, and the model can be used to prioritize high-risk areas for increased enforcement and monitoring.Overall, this study highlights the potential of machine learning models to contribute to improved road safety and reduced accident severity. Further research can be conducted to improve the accuracy and effectiveness of the models, and to explore the use of other variables and data sources for predicting accident severity.

International Journal on Cybernetics & Informatics (IJCI) Vol.12, No.6, December 2023

## AUTHORS

**Adekunle Adefabi** has a BSc in Statistics, double MSc degrees in Statistics and Computer Science. He has over 3 years working experience that cut across Data Management and Software engineering. A researcher and software Engineer with interest in Machine Learning, Data Visualization and Web Backend development.

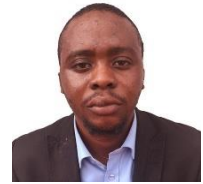

**Somtobe Olisah** is a geoscientist and data analyst with a bachelor's degree in Geological Sciences and double master's degrees in GeoMining, and Predictive Analytics. He has gained experience in geological modeling,and machine learning. He is interested in the application of data analytics across geosciences and believes it has the potential to revolutionize risk assessmentand environmental remediation.

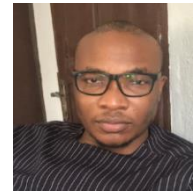

**Callistus Obunadike** holds three Master of Science degrees in geology, mining engineering, and computer science. Callistus combines his extensive knowledge of geosciences with data science. Callistus has a passion for applying machine learning algorithms to improve geological processes and predicting of future event

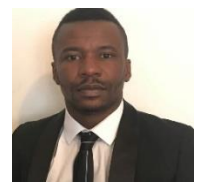

**Oluwatosin Oyetubo**worked on roughly 15 software engineering and machine learning projects and has about 2.5 years of experience in this field. She is dedicated to delivering and utilizing machine learning, business intelligence, and artificial intelligence knowledge to drive business. She holds an MSc in Computer Science in addition to a BSc in Mathematics and Statistics.

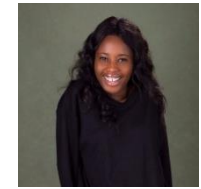

**Esther Taiwo**, a dedicated professional in the IT and data field holds a Bachelor of Science (BSc.) degree in Mathematics and Statistics. Currently, she is pursuing a Master of Science (MSc.) degree in Computer Science and Quantitative Methods. Esther's passion lies in utilizing data to foster innovation and enhance solutions that positively impact our world.

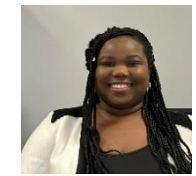

**Edward Tella** is a graduate student of Predictive Analytics at Austin Peay State University, Tennessee and holds a BSc in Systems Engineering from the University of Lagos, Nigeria. Edward has a wealth of professional experience as a Product Manager at two of the largest banks in Africa. His commitment to advancing the field of predictive analytics, coupled with his proven track record of success, makes him poised to drive positive change in the realm of data analytics and make a lasting impact in the technology sector.

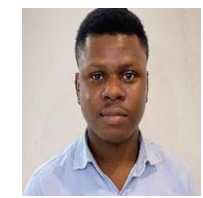